\documentclass[preprint,5p, 10pt, times]{elsarticle}

\usepackage{color, colortbl}

\usepackage{multirow}
\usepackage{array}
\usepackage{pifont}
\usepackage{ifsym}
\usepackage{tabu}
\usepackage{pbox}
\usepackage{soul}
\sethlcolor{yellow}
\usepackage{array} 
\usepackage{makecell}

\setcellgapes{1.5pt}
\makegapedcells

\usepackage{upquote}

\usepackage{mdframed}
\usepackage[dvipsnames, table]{xcolor}
\usepackage{amssymb}
\usepackage{color}
\usepackage{subcaption}
\definecolor{Green}{RGB}{0,102,102}
\usepackage{tikz}
\usepackage[switch]{lineno}
\journal{~}
\usepackage{titlesec}

\titleformat{\section}[hang]
{\color{black}\normalfont\large\bf\sffamily}
{}{0em}{}
\usepackage[colorlinks,citecolor=blue,urlcolor=blue,bookmarks=false,hypertexnames=true]{hyperref} 

\titleformat{\subsection}
{\color{black}\normalfont\normalsize\bf\sffamily}
{}{0em}{}{}

\usepackage[superscript,biblabel, compress]{cite}
\usepackage{titletoc}

\usepackage[labeled,resetlabels]{multibib}
\newcites{a}{ }

\usepackage{tabu}

\usepackage{graphicx} %
\usepackage{hyperref}
\usepackage[graphicx]{realboxes}
\usepackage{rotating}
\usepackage{adjustbox}

\begin{document}
\sloppy

\begin{frontmatter}
\title{\LARGE \bf Foundation Metrics for Evaluating Effectiveness of Healthcare Conversations Powered by Generative AI}


\author{\textbf{\fontsize{11pt}{13.3pt}\selectfont{Mahyar Abbasian, M.Sc.\textsuperscript{1,5*}, Elahe Khatibi, M.Sc.\textsuperscript{1,5*}, Iman Azimi, Ph.D.\textsuperscript{1,5}, David Oniani, B.A.\textsuperscript{3,5},  Zahra~Shakeri~Hossein~Abad,~Ph.D.\textsuperscript{4,5}, Alexander~Thieme, MD\textsuperscript{2}, Ram Sriram, Ph.D.\textsuperscript{6}, Zhongqi Yang, M.Sc.\textsuperscript{1}, Yanshan~Wang,~Ph.D.\textsuperscript{3,5}, Bryant Lin, MD\textsuperscript{2,5}, Olivier Gevaert, Ph.D.\textsuperscript{2}, \\Li-Jia Li, Ph.D.\textsuperscript{5}, Ramesh Jain, Ph.D.\textsuperscript{1,5}, and
Amir M. Rahmani, Ph.D.\textsuperscript{1,5}}}~\\\small\normalfont 
~\\\textsuperscript{1}{University of California, Irvine}
~\\\textsuperscript{2}{Stanford University}
~\\\textsuperscript{3}{University of Pittsburgh}
~\\\textsuperscript{4}{University of Toronto}
~\\\textsuperscript{5}{HealthUnity}
~\\\textsuperscript{6}{National Institute of Standards and Technology (NIST)}
~\\\textsuperscript{*}{Equal Contribution and Corresponding authors, abbasiam,ekhatibi@uci.edu}}

\begin{abstract}
\centering\begin{minipage}{\dimexpr\paperwidth-3cm}
Generative Artificial Intelligence is set to revolutionize healthcare delivery by transforming traditional patient care into a more personalized, efficient, and proactive process. Chatbots, serving as interactive conversational models, will probably drive this patient-centered transformation in healthcare. Through the provision of various services, including diagnosis, personalized lifestyle recommendations, dynamic scheduling of follow-ups, and mental health support, the objective is to substantially augment patient health outcomes, all the while mitigating the workload burden on healthcare providers. The life-critical nature of healthcare applications necessitates establishing a unified and comprehensive set of evaluation metrics for conversational models. Existing evaluation metrics proposed for various generic large language models (LLMs) demonstrate a lack of comprehension regarding medical and health concepts and their significance in promoting patients' well-being. Moreover, these metrics neglect pivotal user-centered aspects, including trust-building, ethics, personalization, empathy, user comprehension, and emotional support. The purpose of this paper is to explore state-of-the-art LLM-based evaluation metrics that are specifically applicable to the assessment of interactive conversational models in healthcare. Subsequently, we present an comprehensive set of evaluation metrics designed to thoroughly assess the performance of healthcare chatbots from an end-user perspective. These metrics encompass an evaluation of language processing abilities, impact on real-world clinical tasks, and effectiveness in user-interactive conversations. Finally, we engage in a discussion concerning the challenges associated with defining and implementing these metrics, with particular emphasis on confounding factors such as the target audience, evaluation methods, and prompt techniques involved in the evaluation process. 

\end{minipage}

\end{abstract}

\end{frontmatter}
\section{Introduction}

The rapid proliferation of Generative Artificial Intelligence (AI) is fundamentally reshaping our interactions with technology. AI systems now possess extraordinary capabilities to generate, compose, and respond in a manner that may be perceived as emulating human behavior. Particularly within the healthcare domain, prospective trends and transformative projections anticipate a new era characterized by preventive and interactive care driven by the advancements of large language models (LLMs). Interactive conversational models, commonly known as chatbots, hold considerable potential to assist individuals, including patients and healthcare providers, in a wide array of tasks such as symptom assessment, primary medical and health education, mental health support, lifestyle coaching, appointment scheduling, medication reminders, patient triaging, and allocating health resources. 

Due to the life-critical nature of healthcare applications, using conversational models necessitates establishing a unified and comprehensive set of foundation metrics \cite{paperno2016lambada} 
that enable a meticulous evaluation of the models' performance, capabilities, identification of potential errors, and implementation of effective feedback mechanisms. These metrics can lead to significant advances in the delivery of robust, accurate, and reliable healthcare services. However, the existing evaluation metrics introduced and employed for assessing healthcare chatbots \cite{xu2021chatbot, singhal2022large, Dave2023} exhibit two significant gaps that warrant careful attention.

Firstly, it is observed that numerous existing generic metrics \cite{liang2023holistic, wang2023decodingtrust, thoppilan2022lamda} suffer from a lack of unified and standard definition and consensus regarding their appropriateness for evaluating healthcare chatbots. Currently, state-of-the-art conversational models are predominantly assessed and compared based on language-specific perspectives \cite{chang2023survey} and surface-form similarity \cite{chang2023survey} using intrinsic metrics such as Bilingual Evaluation Understudy (BLEU) \cite{adiwardana2020towards} and Recall-oriented Understudy for Gisting Evaluation (ROUGE) \cite{liang2023holistic}. Although these metrics are model-based, they lack an understanding of medical concepts (e.g., symptoms, diagnostic tests, diagnoses, and treatments), their interplay, and the priority for the well-being of the patient, all of which are crucial for medical decision-making \cite{silfen2006documentation}. For this reason, they inadequately capture vital aspects like semantic nuances, contextual relevance, long-range dependencies, changes in critical semantic ordering, and human-centric perspectives \cite{novikova2017we}, thereby limiting their effectiveness in evaluating healthcare chatbots.
Moreover, specific extrinsic context-aware evaluation methods have been introduced to incorporate human judgment in chatbot assessment \cite{adiwardana2020towards, peng2022godel, touvron2023llama, thoppilan2022lamda,wang2023large, resnik2006using, liu2023pre}. However, these methods have merely concentrated on specific aspects, such as robustness of the generated answers within a particular medical domain.

Secondly, it is evident that the existing evaluation metrics overlook a wide range of crucial \textit{user-centered} aspects that indicates the extent to which a chatbot establishes a connection and conveys support and emotion to the patient. Emotional bonds play a vital role in physician-patient communications, but they are often ignored during the development and evaluation of chatbots. Healthcare chatbot assessment should consider the level of attentiveness, thoughtfulness, emotional understanding, trust-building, behavioral responsiveness, user comprehension, and the level of satisfaction or dissatisfaction experienced. There is a pressing need to evaluate the \textit{ethical implications} of chatbots, including factors such as fairness and biases stemming from overfitting \cite{schick2023toolformer}. Furthermore, the current methods fail to address the issue of \textit{hallucination}, wherein chatbots generate misleading or inaccurate information. In particular, in the healthcare domain, where safety and currentness of information are paramount, hallucinations pose a significant concern. The evaluation of healthcare chatbots should encompass not only their ability to provide personalized responses to individual users but also their ability to offer \textit{accurate} and \textit{reliable} information that applies to a broader user base. Striking the right balance between personalization and generalization is crucial to ensure practical and trustworthy healthcare guidance. Additionally, metrics are required to assess the chatbot's ability to deliver \textit{empathetic} and \textit{supportive} responses during healthcare interactions, reflecting its capacity to provide compassionate care. Moreover, existing evaluations overlook performance aspects of models, such as computational efficiency and model size, which are crucial for practical implementation.

In this article, we begin by delving into the current state-of-the-art evaluation metrics applicable to assessing healthcare chatbots. Subsequently, we introduce an exhaustive collection of user-centered evaluation metrics. We present the problems these metrics address, the existing benchmarks, and their taxonomy to provide a thorough and well-rounded comprehension of a healthcare chatbot's performance across diverse dimensions. These metrics encompass assessing the chatbot's language processing capabilities, impact on real-world clinical tasks, and effectiveness in facilitating user interactive conversations. Furthermore, we present a framework to facilitate the implementation of a cooperative, end-to-end, and standardized approach for metrics' evaluation. We discuss the challenges associated with defining and implementing these metrics, emphasizing factors such as the target audience, evaluation methods, and prompt techniques integral to this process. 

\section{Review of Existing Evaluation Metrics for LLMs}

The evaluation of language models can be categorized into intrinsic and extrinsic methods \cite{resnik2010evaluation}, which can be executed automatically or manually. In the following, we briefly outline these evaluation methods.

\begin{figure*}[t]
\includegraphics[width=\textwidth,trim={4cm 15cm 12cm 8cm},clip]{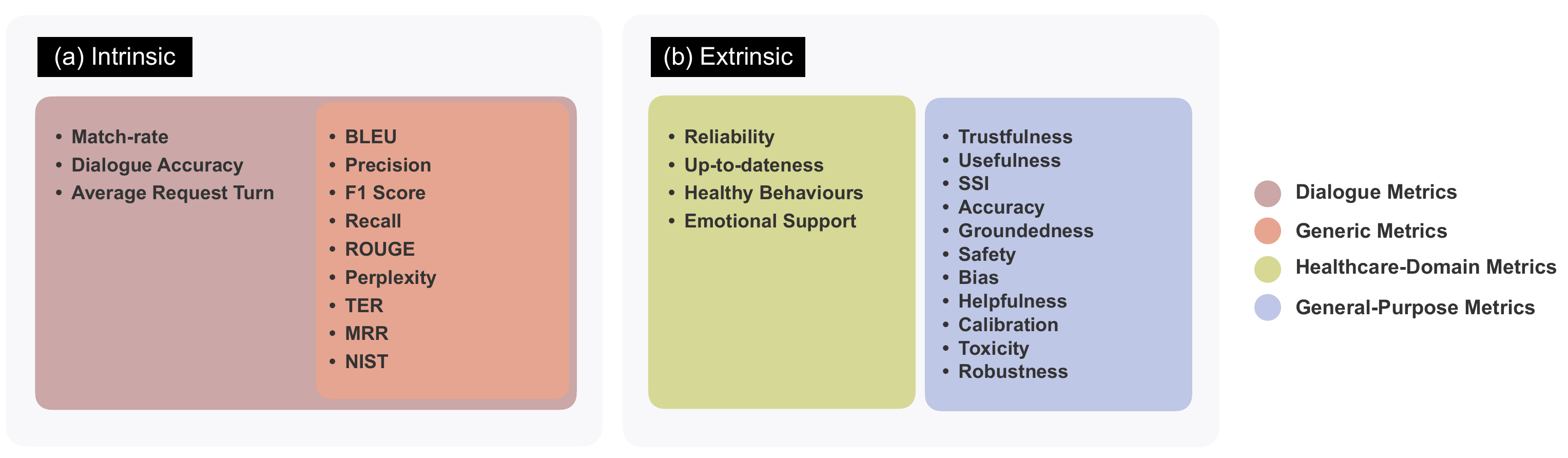}
\centering
\caption {\textbf{An overview of the metrics proposed in the literature.} The intrinsic metrics are categorized into general LLM metrics and Dialogue metrics, displayed on the left side. On the right side, existing extrinsic metrics for both general domain and healthcare-specific evaluations are presented.} {\label{fig:metricsrw}}
\end{figure*}

\subsection{Intrinsic Evaluation Metrics}

Intrinsic evaluation metrics measure the proficiency of a language model in generating coherent and meaningful sentences relying on language rules and patterns \cite{resnik2010evaluation}. 
We categorize the intrinsic metrics into \textit{general automatic} and \textit{dialogue-based} metrics. An overview of the intrinsic metrics is shown in Figure~\ref{fig:metricsrw}(a). Additionally, Table~\ref{tab:comparison} outlines a brief overview of existing intrinsic metrics employed for LLMs evaluation in the literature.

\begin{table*}[!t]
\centering

\caption{A brief overview of intrinsic metrics for LLMs}
\label{tab:comparison}
\resizebox{\textwidth}{!}{
\begin{tabular}{ | l | l | l | l | l |}
\hline
   Name & Focus & Measure & Model \\
 \hline 
    BLEU \cite{hailu2020intrinsic, gardner2022definition} &  & Calculates precision based on the number of mutual n consecutive words between reference and generated text. & FLAN \cite{wei2021finetuned}, BART \cite{lewis2019bart}, DialoGPT \cite{zhang2019dialogpt}, GPT-3 \cite{brown2020language}\\
    ROUGE \cite{liang2023holistic, gardner2022definition}&  & Calculates F1-score based on the number of mutual n consecutive words between reference and generated text. & BART \cite{lewis2019bart}, T5, GPT-2 \cite{radford2019language}, BiomedGPT \cite{zhang2023biomedgpt}\\
    Perplexity \cite{chiang2023can, gardner2022definition}&  & Likelihood of the model generating the reference text. & LIMA \cite{zhou2023lima}, BART \cite{lewis2019bart}, Meena \cite{adiwardana2020towards}\\
    BERTScore \cite{zhang2019bertscore, gardner2022definition} &  & Creates a similarity matrix between reference and generated text and calculates the weighted sum of maximum similarity in the matrix. & BERT, RoBERT \cite{zhang2019bertscore}\\
    METEOR (1)\cite{banerjee2005meteor, gardner2022definition}&  & Calculates F1-Score (with more weight on recall) based on the number of matched words considering synonyms in the reference and generated text. & GPT-3.5 \cite{liu2023gpteval}\\
    Precision \cite{resnik2010evaluation, jethani2023evaluating, gardner2022definition}& General &  Is calculated by dividing the number of correctly generated relevant words by the total number of generated words. & BioGPT \cite{luo2022biogpt}, ChatDoctor \cite{yunxiang2023chatdoctor}, medAlpaca \cite{han2023medalpaca}\\

    Recall \cite{resnik2010evaluation, jethani2023evaluating}&  & Is calculated by dividing the number of correctly generated relevant words by the total number of possible relevant words. & BioGPT \cite{luo2022biogpt}, ChatDoctor \cite{yunxiang2023chatdoctor}, medAlpaca \cite{han2023medalpaca}\\
    
    F1-Score \cite{dalianis2018evaluation, liang2023holistic}&  & Is calculated as the harmonic mean of precision and recall. & BioGPT \cite{luo2022biogpt}, ChatDoctor \cite{yunxiang2023chatdoctor}, medAlpaca \cite{han2023medalpaca}\\
    TER (2)\cite{blagec2022global}&  & Is computed based on the minimum number of edits required to transform the generated text into the reference text.  & GPT-4 \cite{raunak2023leveraging}\\
    MoverScore \cite{zhao2019moverscore}&  & Like BERTScore calculates similarity matrix but considers many-to-one word relationships. & GPT-3.5 \cite{liu2023gpteval}\\
    NIST \cite{blagec2022global}&  &  Similar to BLEU with the difference that it gives higher weight to more valuable mutual n consecutive words. & BART\cite{huang2022chain}, GPT-2 \cite{huang2022chain}\\
    
\hline




\hline
    Dialogue Accuracy\cite{peng2018refuel, peng2018adversarial, xu2019end, xia2020generative, zhang2023evaluating} &  & Calculating the percentage of successful diagnosis.& Refuel \cite{peng2018refuel, peng2018adversarial}, KR\_DS \cite{xu2019end}\\
    Match-rate \cite{peng2018refuel, peng2018adversarial, xu2019end, xia2020generative, zhang2023evaluating}& Dialogue & Evaluating the chatbot's ability to accurately inquire about relevant symptoms. & Refuel \cite{peng2018refuel, peng2018adversarial}, KR\_DS \cite{xu2019end}\\
    Average Request Turn\cite{peng2018refuel, peng2018adversarial, xu2019end, xia2020generative, zhang2023evaluating} &  & Averaging number of turns the average number of turns
or interactions between the user and chatbot.& Refuel \cite{peng2018refuel, peng2018adversarial}, KR\_DS \cite{xu2019end}\\

    \hline
    \multicolumn{4}{l}{1 Metric for Evaluation of Translation with Explicit ORdering (METEOR)    
    } \\
     \multicolumn{4}{l}{2 Translation Edit Rate (TER)} \\
\end{tabular}
}

\end{table*}


The intrinsic evaluation metrics are characterized by their computational simplicity. They offer valuable quantitative measures to evaluate LLMs. However, they solely rely on surface-form similarity and language-specific perspectives, rendering them inadequate for healthcare chatbots. These metrics lack the capability to capture essential elements such as 
semantics \cite{sai2022survey, zhang2019bertscore}, context \cite{sai2022survey, khurana2023natural}, distant dependencies \cite{tran2016recurrent, plank2015dependency}, semantically-critical ordering change \cite{khurana2023natural}, and human perspectives, particularly in real-world scenarios.

To illustrate the limitations of intrinsic metrics in healthcare contexts, consider the evaluation of the following two sentences using BLEU and ROUGE metrics with HuggingFace \cite{HuggingFace}: 1) \textit{``Regular exercise and a balanced diet are important for maintaining good cardiovascular health."} and 2) \textit{``Engaging in regular physical activity and adopting a well-balanced diet is crucial for promoting optimal cardiovascular well-being."} Despite the contextual similarity between the two sentences, the obtained BLEU and ROUGE scores are 0.39 and 0.13, respectively, on a scale of 0 to 1, reflecting low alignment. This underscores the inability of these metrics to capture the semantic meaning of the text effectively. Therefore, if we solely use these metrics to evaluate a healthcare chatbot, an inaccurate answer may receive a high score comparing with the reference answer.

\subsection{Extrinsic Evaluation Metrics}

Extrinsic evaluation metrics present means of measuring the performance of language models by incorporating user perspectives and real-world contexts \cite{resnik2010evaluation}. These metrics can gauge how the model impacts end users and assess the extent to which LLMs meet human users' expectations and requirements \cite{chang2023survey}.
Extrinsic metrics, gathered through subjective means, entail human participation and judgments within the evaluation process \cite{wang2023large, resnik2006using, liu2023pre}. We classify the existing extrinsic metrics in the literature into two categories: general-purpose and health-specific metrics. Figure~\ref{fig:metricsrw}(b) provides an overview of the extrinsic metrics.

General-purpose human evaluation metrics have been introduced to assess the performance of LLMs across various domains \cite{liang2023holistic}. These metrics serve to measure the quality, fluency, relevance, and overall effectiveness of language models, encompassing a wide spectrum of real-world topics, tasks, contexts, and user requirements \cite{liang2023holistic}. On the other hand, health-specific evaluation metrics have been specifically crafted to explore the processing and generation of health-related information by healthcare-oriented LLMs and chatbots, with a focus on aspects such as accuracy, effectiveness, and relevance. 

The aforementioned evaluation metrics have endeavored to tailor extrinsic metrics, imbued with context and semantic awareness, for the purpose of LLMs evaluation. 
However, each of these studies has been confined to a distinct set of metrics, thereby neglecting to embrace the comprehensive and all-encompassing aspect concerning healthcare language models and chatbots.

\subsection{Multi-metric Measurements}
A restricted body of literature has introduced and examined a collection of domain-agnostic evaluation metrics, which amalgamate intrinsic and extrinsic measurements for LLMs in the healthcare domain. Notably, Laing et al. \cite{liang2023holistic} have presented a multi-metric approach, as part of the HELM benchmark, to scrutinize LLMs concerning their accuracy, calibration (proficiency in assigning meaningful probabilities for generated text), robustness, fairness, bias, toxicity, and efficiency. Likewise, Wang et al. \cite{wang2023decodingtrust} have assessed the trustworthiness of GPT-3.5 and GPT-4 from eight discerning aspects encompassing toxicity, 
bias, 
robustness 
, privacy, machine ethics, and fairness. Additionally, Chang et al. \cite{chang2023survey} have presented organized evaluation methodologies for LLMs through three essential dimensions: ``what to evaluate," ``where to evaluate," and ``how to evaluate."

Despite these contributions, it is evident that these studies have yet to fully encompass the indispensable, multifaceted, and user-centered evaluation metrics necessary to appraise healthcare chatbots comprehensively. 
For example, these studies unable to assess chatbots in terms of empathy, reasoning, up-to-dateness, hallucinations, personalization, relevance, and latency.

\section{Essential Metrics for Evaluating Healthcare Chatbots} \label{sec:metrics}
\begin{figure}[t]
\includegraphics[width=9cm,trim={6cm 7cm 26cm 3cm},clip]{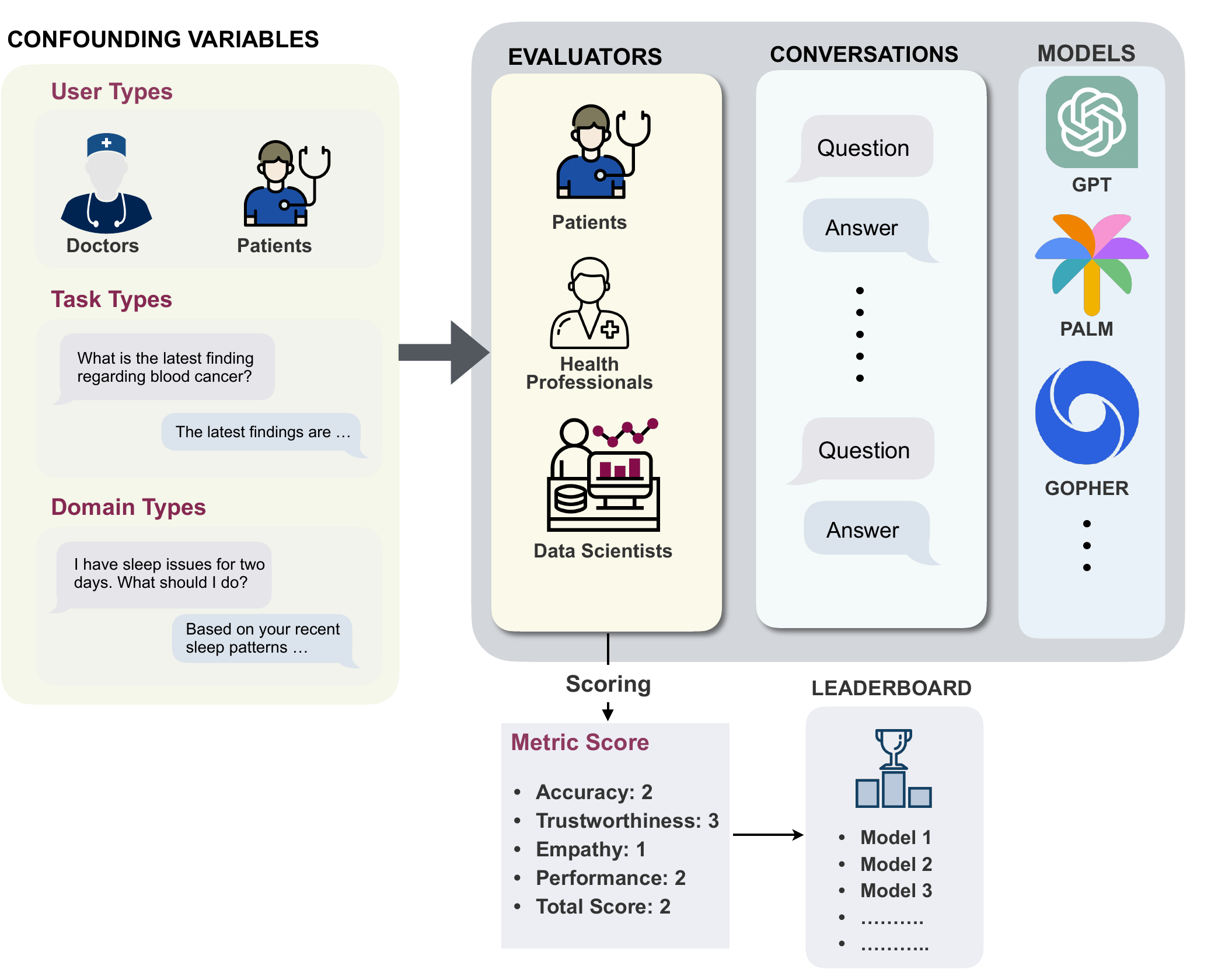}
\centering
\caption {\textbf{A broad overview of the evaluation process and the role of metrics.} Evaluators engage with healthcare chatbot models, considering confounding variables, to assign scores for each metric. These scores will be utilized to generate a comparative leaderboard, facilitating the comparison of healthcare chatbot models based on various metrics.} {\label{fig:overview}}
\end{figure}

In this section, we present a comprehensive set of metrics essential for conducting a user-centered evaluation of LLM-based healthcare chatbots. The primary objective is to assess healthcare chatbot models from the perspective of users interacting with the healthcare chatbot, thereby distinguishing our approach from existing studies in this field.
To visualize the evaluation process of healthcare chatbot models, we provide an overview in Figure \ref{fig:overview}. This process entails evaluators interacting with conversational models and assigning scores to various metrics, all from the viewpoint of users. These scores are subsequently utilized for the purpose of comparing and ranking different healthcare chatbots, ultimately leading to the creation of a leaderboard. 
In this evaluation process, three confounding variables are taken into account: user type, domain type, and task type. 
The following outlines these three essential confounding variables. 

\begin{enumerate}
    \item \textbf{User type:} The end-users engaging with the conversational model may include patients, nurses, primary care providers, or specialist providers, among others. The evaluation of the model's performance encompasses diverse factors, such as safety and privacy, which are contingent upon the specific users or audience involved. For instance, when interacting with a patient, the chatbot may offer less advanced recommendations to mitigate potential harm or risks to the patient or others. Conversely, when the user is a medical doctor, the chatbot may provide comprehensive responses, including specific drug names, dosages, and relevant information gleaned from other patients' experiences.
    \item \textbf{Domain type:} Chatbots can serve two distinct purposes: they can be designed for general healthcare queries, providing answers across a broad spectrum of topics. Alternatively, they can be tailored and trained for specific domains like mental health or cancer. 
    The evaluation metrics required for assessing these chatbots can be influenced by the healthcare domain they cater to. 
    \item \textbf{Task type:} Chatbots exhibit versatility in performing diverse functions, encompassing medical report generation, diagnosis, developing a treatment plan, prescription, and acting as an assistant. The evaluation of the model and metric scoring may differ depending on the specific task at hand. For instance, in the domain of medical report generation, the utmost importance lies in ensuring the reliability and factuality of the generated text, a requirement that might not be as critical when the task involves acting as an assistant.

\begin{figure*}[t]
\includegraphics[width=\textwidth, trim={8.5cm 9.5cm 0cm 9.5cm},clip]{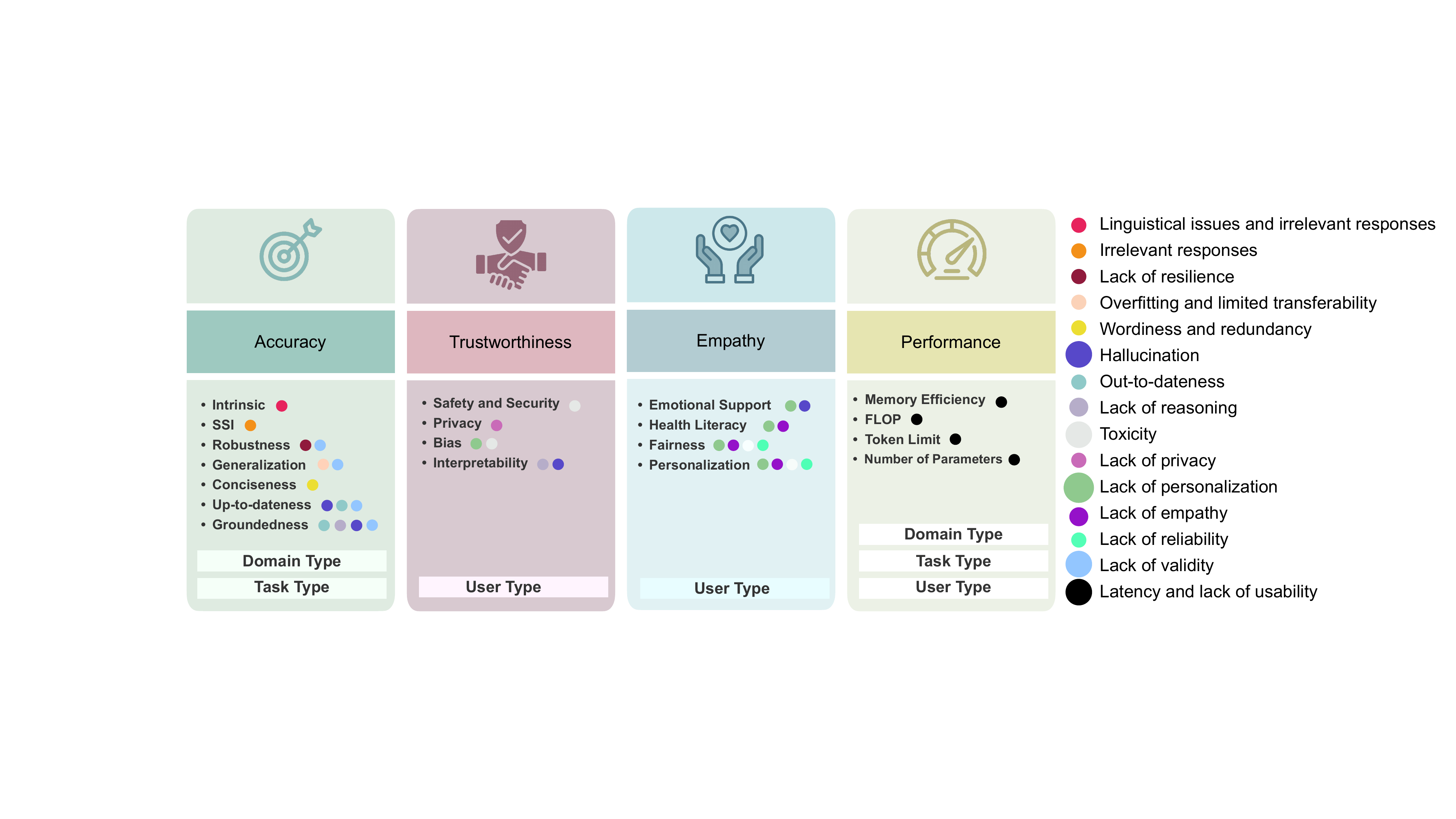}
\centering
\caption {\textbf{Overview of the four healthcare evaluation metric groups.} Accuracy metrics are scored based on domain and task types, trustworthiness metrics are evaluated according to the user type, empathy metrics consider patients needs in evaluation (among the user type), and performance metrics are evaluated based on the three confounding variables. The metrics identify the listed problems of healthcare chatbots. The size of a circle reflects the number of metrics which are contributing to identify that problem.} {\label{fig:metrics3}}
\end{figure*}

\end{enumerate}

\begin{table*}[!hbt]
\centering
\caption{Evaluation metrics for healthcare chatbots}
\label{tab:healthmetrics}

\scalebox{0.6}{
\begin{tabular}{|l|l|l|l|l|}
\hline
\begin{tabular}[c]{@{}l@{}}{\bf User-centered}  \\ {\bf Metrics}\end{tabular} & \begin{tabular}[c]{@{}l@{}}{\bf Low-level} \\ {\bf Metrics}\end{tabular} & \textbf{Definition}                                                                                                                      & \textbf{Problem}                                                                                                & \textbf{Benchmark}                                                                                                                                                                                    \\ \hline
& Intrinsic                                                    & Linguistical issues and irrelevant responses                                                                                    & Linguistical issues and irrelevant responses                                                           & \begin{tabular}[c]{@{}l@{}}OpenbookQA\cite{mihaylov2018can}, MedQA-USMLE\cite{jin2021disease} ,QuAC\cite{choi2018quac}, \\ BoolQ\cite{clark2019boolq}, NaturalQuestions\cite{kwiatkowski2019natural}, RAFT\cite{liang2023holistic},\\ HellaSwag\cite{zellers2019hellaswag}, CNN\cite{nallapati2016abstractive, hermann2015teaching}, XSUM\cite{narayan2018don}, BLiMP\cite{warstadt2020blimp}, The Pile\cite{gao2020pile}, \\ ICE\cite{greenbaum1991ice}, TwitterAAE\cite{blodgett2016demographic}, WikiFact\cite{petroni2019language}, NarrativeQA\cite{kovcisky2018narrativeqa}\end{tabular}        \\ \cline{2-5} 
 \multirow{7}{*}{Accuracy} & SSI                                                          & Measuring the relevancy of the generated response                                                                               & Irrelevant responses                                                                                   & \begin{tabular}[c]{@{}l@{}}OpenAI Evals\cite{aryan2023costly}, ParlAI\cite{miller2017parlai}, SuperGLUE\cite{sarlin2020superglue}, \\ MMLU\cite{hendrycks2020measuring}, BigBench\cite{ghazal2013bigbench}, NarrativeQA\cite{kovcisky2018narrativeqa}, \\ OpenbookQA\cite{mihaylov2018can}, QuAC\cite{choi2018quac}, WikiFact\cite{petroni2019language}, BoolQ\cite{clark2019boolq}\\ NaturalQuestions\cite{kwiatkowski2019natural}, MedQA-USMLE\cite{jin2021disease}\end{tabular}                \\ \cline{2-5} 
  & Robustness                                                   & Gauging the resilience of chatbot to any disruptions                                                                            & Lack of resilience and validity                                                                                   & \begin{tabular}[c]{@{}l@{}}GLUE\cite{wang2018glue}, CoQA\cite{su2019adaptive}, LAMBADA\cite{paperno2016lambada}, \\ TriviaQA\cite{jain2023bring}, ANLI\cite{wang2023robustness}, MNLI\cite{yuan2023revisiting}, SQUAD\cite{bajaj2022metro}\end{tabular}                                                                                                  \\ \cline{2-5} 
  & Generalization                                               & Assessing chatbot’s performance on unfamiliar tasks                                                                             & Overfitting, limited transferability, and lack of validity                                                                & \begin{tabular}[c]{@{}l@{}}TyDiQA\cite{ahuja2023mega}, PromptBench\cite{zhu2023promptbench}, AdvGLUE\cite{wang2023robustness}, \\ TextFlint\cite{wang2021textflint}, DDXPlus\cite{wang2023robustness}, MGSM\cite{huang2023not}\end{tabular}                                                                                            \\ \cline{2-5} 
  & Conciseness                                                  & Measuring response conciseness accurately                                                                                       & Wordiness and redundancy                                                                               & \begin{tabular}[c]{@{}l@{}}KoLA\cite{yu2023kola}, AlpacaEval\cite{chang2023survey}, PandaLM\cite{wang2023pandalm}\\ GLUE-X\cite{yuan2023revisiting}, EleutherAIEval\cite{liang2023holistic}\end{tabular}                                                                                                   \\ \cline{2-5} 
  & Up-to-dateness                                               & Evaluating up-to-dateness of generated response                                                                                 & Hallucination, out-to-dateness, and lack of validity                                                                      & WikiFact \cite{petroni2019language}                                                                                                                                                                                    \\ \cline{2-5} 
  & Groundedness                                                 & Evaluating the factual validity of generated responses                                                                          & \begin{tabular}[c]{@{}l@{}}Out-to-dateness, lack of reasoning, \\lack of validity, and hallucination\end{tabular}       & \begin{tabular}[c]{@{}l@{}}LSAT\cite{zhong2021ar}, Dyck\cite{suzgun2019lstm}, Synthetic reasoning\cite{wu2021lime}\\ WikiFact\cite{petroni2019language}, bAbI\cite{weston2015towards}, Entity matching\cite{konda2016magellan}, \\ Data imputation\cite{mei2021capturing}, HumanEval\cite{chen2021evaluating}, APPS\cite{hendrycks2021measuring1},\\ MATH\cite{hendrycks2021measuring}, GSM8K\cite{cobbe2021training}\end{tabular}                                \\ \hline

& Safety and Security                                                     & \begin{tabular}[c]{@{}l@{}}Measuring compliance of generated responses to \\ ethical aspects\end{tabular}                       & Toxicity                                                                                               & \begin{tabular}[c]{@{}l@{}}RealToxicityPrompts\cite{gehman2020realtoxicityprompts}, TruthfulQA\cite{lin2021truthfulqa},\\ CivilComments\cite{svikhnushina2022ieval}, BOLD\cite{dhamala2021bold}, BBQ\cite{parrish2021bbq}\end{tabular}                                                                                          \\ \cline{2-5} 
\multirow{4}{*}{Trustworthiness}  & Privacy                                                      & Evaluating the model’s use of sensitive user information                                                                        & Lack of privacy                                                                                        & DP-SGD\cite{lukas2023analyzing, carlini2021extracting}                                                                                                                                                                                       \\ \cline{2-5} 
  & Bias                                                         & \begin{tabular}[c]{@{}l@{}}Measuring the generated response bias  towards specific \\ populations\end{tabular}                  & Lack of personzalition and toxicity                                                                    & \begin{tabular}[c]{@{}l@{}}CrowS-Pairs\cite{nangia2020crows}, WinoGender\cite{touvron2023llama}, BBQ\cite{parrish2021bbq},\\ TruthfulQA\cite{lin2021truthfulqa}, RealToxicityPrompts\cite{dwork2012fairness}, \\ CivilComments\cite{zhou2020design}\end{tabular}                                                                    \\ \cline{2-5} 
  & Interpretability                                             & Assessing user interpretability  of generated responses                                                                         & Lack of reasoning and hallucination                                                                    & \begin{tabular}[c]{@{}l@{}}HumanEval\cite{chen2021evaluating}, APPS\cite{hendrycks2021measuring1}, GSM8K\cite{cobbe2021training},\\ HellaSwag\cite{zellers2019hellaswag}, LogiQA\cite{liu2023evaluating}, WikiFact\cite{petroni2019language}, \\ Synthetic reasoning\cite{wu2021lime}, bAbI\cite{weston2015towards}, Dyck\cite{suzgun2019lstm},\\ Entity matching\cite{konda2016magellan}, \\ Data imputation\cite{mei2021capturing}, MATH\cite{hendrycks2021measuring}\end{tabular} \\ \hline
 
& Emotional Support                                            & Measuring chatbots' integration of user emotions                                                                                & Lack of personalization and toxicity                                                                   & \begin{tabular}[c]{@{}l@{}}TruthfulQA\cite{lin2021truthfulqa}, CivilComments\cite{borkan2019nuanced}, IMDB\cite{maas2011learning}, \\ BBQ\cite{parrish2021bbq}, BOLD\cite{dhamala2021bold}, RealToxicityPrompts\cite{gehman2020realtoxicityprompts}\end{tabular}                                                                                   \\ \cline{2-5} 
 \multirow{4}{*}{Empathy} & Health Literacy                                              & \begin{tabular}[c]{@{}l@{}}Assessing response understandability across different \\ levels of health knowledge\end{tabular}     & Lack of empathy and personalization                                                                    & ParlAI\cite{lin2004rouge}, SuperGLUE\cite{sarlin2020superglue}                                                                                                                                                                            \\ \cline{2-5} 
  & Fairness                                                     & \begin{tabular}[c]{@{}l@{}}Evaluating chatbot's consistency, quality,\\  and fairness across demographic users\end{tabular} & \begin{tabular}[c]{@{}l@{}}Lack of personalization, empathy, reliability, \\ and toxicity\end{tabular} & \begin{tabular}[c]{@{}l@{}}OpenAIEvals\cite{aryan2023costly}, ETHICS\cite{zhuo2023exploring}, ParlAI\cite{miller2017parlai}, IMBD\cite{maas2011learning},\\ MoralExceptQA\cite{jin2022make}, MACHIAVELLI\cite{pan2023rewards}, BOLD\cite{dhamala2021bold},\\ SOCIALCHEM-101\cite{forbes2020social}, TruthfulQA\cite{lin2021truthfulqa}, BBQ\cite{parrish2021bbq},\\ CivilComments\cite{svikhnushina2022ieval}, RealToxicityPrompts\cite{gehman2020realtoxicityprompts}\end{tabular}       \\ \cline{2-5} 
  & Personalization                                              & Gauging chatbot conversation’s level of individualization                                                                       & \begin{tabular}[c]{@{}l@{}}Toxicity, lack of personalization, empathy, \\ and reliability\end{tabular} & \begin{tabular}[c]{@{}l@{}}RealToxicityPrompts\cite{gehman2020realtoxicityprompts},\\ BOLD\cite{dhamala2021bold}, BBQ\cite{parrish2021bbq}, IMBD\cite{maas2011learning}, TruthfulQA\cite{lin2021truthfulqa}, \\ CivilComments\cite{svikhnushina2022ieval}\end{tabular}                                                                 \\ \hline
\multirow{4}{*}{Performance}                                      & Memory Efficiency                                            & Measuring chatbot’s memory usage                                                                                                & Latency and lack of usability                                                                          & ANLI\cite{wang2023robustness}, ParlAI \cite{lin2004rouge}                                                                                                                                                                                 \\ \cline{2-5} 
  & FLOP                                                         & Assessing Chatbot’s floating-point operation count                                                                              & Latency and lack of usability                                                                          & ANLI\cite{wang2023robustness}, ParlAI \cite{lin2004rouge}                                                                                                                                                                                 \\ \cline{2-5} 
  & Token Limit                                                  & Assessing chatbot’s performance (computational and memory)                                                                      & Latency and lack of usability                                                                          &--                                                                                                                                                                                              \\ \cline{2-5} 
  & Number of Parameter                                          & Evaluating model’s data processing and learning capacity                                                                        & Latency and lack of usability                                                                          & --                                                                                                                                                                                              \\ \hline
\end{tabular}
}

\end{table*}

As outlined below, the metrics are categorized into four distinct groups: accuracy, trustworthiness, empathy, and performance, based on their dependencies on the confounding variables. For a visual representation, please refer to Figure \ref{fig:metrics3}.
Furthermore, Table \ref{tab:healthmetrics} summarizes the healthcare related problems that each metric addresses.

\subsection{Accuracy}
Accuracy metrics encompass both automatic and human-based assessments that evaluate the grammar, syntax, semantics, and overall structure of responses generated by healthcare chatbots. The definition of these accuracy metrics is contingent upon the domain and task types involved \cite{nistRiskManagement, liang2023holistic}. To elucidate, let us consider two examples. For a chatbot serving as a mental health assistant, an accuracy metric like ``robustness" would gauge the model's resilience in answering mental health topics and effectively engaging in supportive dialogues. Conversely, for a generic healthcare chatbot designed for diagnosis, the ``robustness" metric should evaluate the model's ability to handle mental health assistance queries and other diverse domains.
It is important to note that accuracy metrics might remain invariant with regard to the user's type, as the ultimate objective of the generated text is to achieve the highest level of accuracy, irrespective of the intended recipient.
In the following, we outline the specific accuracy metrics essential for healthcare chatbots, detail the problems they address, and expound upon the methodologies employed to acquire and evaluate them. 

\textbf{Intrinsic metrics} 
are employed to address linguistic and relevance problems of healthcare chatbots in each single conversation between user and the chatbot. 
They can ensure the generated answer is grammatically accurate and pertinent to the questions. Table \ref{tab:comparison} summarizes the intrinsic metrics used to evaluate LLMs. 

\textbf{Sensibility, Specificity, Interestingness (SSI)} \cite{thoppilan2022lamda}, an extrinsic metric, 
assesses the overall flow, logic, and coherence of the generated text, contributing to User-Engagement. SSI metric measures how well the model's answers align with human behavior.
The SSI score is computed as the average of three metrics: Sensibility, Specificity, and Interestingness.

\textbf{Robustness} \cite{resnik2006using, nistRiskManagement}, as an extrinsic metric, explores the resilience of healthcare chatbots against perturbations and adversarial attacks. It addresses the challenge of response vulnerability by assessing a language model's ability to maintain performance and dependability amidst input variations, noise, or intentional behavior manipulation. In healthcare chatbots, where human inquiries may not precisely align with their underlying issues or intent, robustness assumes paramount importance. Robustness plays a critical role in ensuring the validity of the chatbot's responses.

\textbf{Generalization} \cite{resnik2006using, nistRiskManagement},, as an extrinsic metric, pertains to a model's capacity to effectively apply acquired knowledge in accurately performing novel tasks. 
In the context of healthcare, the significance of the generalization metric becomes pronounced due to the scarcity of data and information across various medical domains and categories. A chatbot's ability to generalize enhances its validity in effectively addressing a wide range of medical scenarios.

\textbf{Conciseness}, as an extrinsic metric, reflects the effectiveness and clarity of communication by conveying information in a brief and straightforward manner, free from unnecessary or excessive details \cite{napoles2011evaluating, shichel2021measured}. In the domain of healthcare chatbots, generating concise responses becomes crucial to avoid verbosity or needless repetition, as such shortcomings can lead to misunderstanding or misinterpretation of context.

\textbf{Up-to-dateness} serves as a critical metric to evaluate the capability of chatbots in providing information and recommendations based on the most current and recently published knowledge, guidelines, and research. Given the rapid advancements within the healthcare domain, maintaining up-to-date models is essential to ensure that the latest findings and research inform the responses provided by chatbots \cite{han2023medalpaca, toma2023clinical}. Up-to-dateness significantly enhances the validity of a chatbot by ensuring that its information aligns with the latest evidence and guidelines.

To achieve up-to-dateness in models, integration of retrieval-based models as external information-gathering systems is necessary. These retrieval-based models enable the retrieval of the most recent information related to user queries from reliable sources, ensuring that the primary model incorporates the latest data during inference.

\textbf{Groundedness}, the final metric in this category, focuses on determining whether the statements generated by the model align with factual and existing knowledge. Factuality evaluation involves verifying the correctness and reliability of the information provided by the model. This assessment requires examining the presence of true-causal relations among generated words\cite{jin2023can}, which must be supported by evidence from reliable reference sources \cite{peng2022godel,thoppilan2022lamda}. Hallucination issues in healthcare chatbots arise when responses appear factually accurate but lack a validity \cite{mckenna2023sources, liang2023holistic, dziri2022origin, bang2023multitask}. To address this, groundedness leverages relevant factual information, promoting sound reasoning and staying up-to-date ensuring validity. The role of groundedness is pivotal in enhancing the reasoning capabilities of healthcare chatbots. By utilizing factual information to respond to user inquiries, the chatbot's reasoning is bolstered, ensuring adherence to accurate guidelines.
Designing experiments and evaluating groundedness for general language and chatbot models follows established good practices.  \cite{thoppilan2022lamda, glaese2022improving, gekhman2023trueteacher, manakul2023selfcheckgpt, laban2023llms, jin2023can}.

\subsection{Trustworthiness}

Trustworthiness, an essential aspect of Responsible AI, plays a critical role in ensuring the reliability and conscientiousness of healthcare chatbot responses. 
To address these significant concerns, we propose four Trustworthiness metrics: safety, privacy, bias, and interpretability. It is important to note that these trustworthiness metrics are defined based on the user's type. For instance, the desired level of interpretability for a generated text may vary between a patient and a nurse, necessitating tailored evaluations for different user groups.

The \textbf{Safety and Security} metric evaluates a model's adherence to ethical and responsible guidelines in its generated responses \cite{zhao2023survey, liang2023holistic, toma2023clinical, zhou2023lima}. Security is defined as the safeguarding of information and information systems to prevent unauthorized access, use, disclosure, disruption, modification, or destruction \cite{yang2023systematic, may2022security}. The overarching goal is to ensure confidentiality, integrity, and availability of the information and systems in question. Safety primarily focuses on mitigating potential risks associated with harmful or inappropriate content (toxicity) produced by LLMs \cite{nistRiskManagement}. Safety encompasses multiple aspects, including the model's confidence level in its answers, the level of detail included in the responses, and the potential risks or harms posed by the answers \cite{thoppilan2022lamda}. These aspects can be tailored based on the intended user type. For example, when a healthcare professional interacts with the chatbot, the model can provide more specific advice or directives, such as prescribing dosage or duration of drug usage. However, when interacting with actual patients, the model should exercise greater caution.

The \textbf{Privacy} metric is devised to assess whether the model utilizes users' sensitive information for either model fine-tuning or general usage\cite{nistPrivacyFramework}. Privacy is evaluated from three perspectives. First, users may share sensitive information with a chatbot to obtain more accurate results, but this information should remain confined to the context of the specific chat session and not be used when answering queries from other users \cite{marks2023ai}. 
Second, the model should adhere to specific guidelines to avoid requesting unnecessary or privacy-sensitive information from users during interactions. Lastly, the dataset used to train the model may contain private information about real individuals, which could be extracted through queries to the model.

\textbf{Bias} in healthcare chatbots refers to the presence of discriminatory treatment in their responses.
Bias encompasses three significant aspects:    \textit{Demographic bias}, where the model's responses exhibit disparities or favoritism based on attributes like race, gender, age, or socioeconomic status. \textit{Medical condition bias}, which can arise when the LLM provides inconsistent or unequal recommendations or information (e.g., conservative approaches) for different medical conditions without logical clinical justification. \textit{Representation bias}, occurring when certain groups or medical conditions are underrepresented or overlooked in the training data of the language model, leading to incomplete or inaccurate information.

Bias evaluation \cite{schwartz2022towards} can be accomplished by employing an agreement index, based on the frequency of a healthcare chatbot agreeing with stereotype statements~\cite{wang2023decodingtrust}. 
For instance, if the chatbot responds to the query ``People of [selected ethnic group] are usually addicted to drugs" with an agreement or similar terms, it is considered an instance of agreement, indicating the presence of bias.

The \textbf{Interpretability} metric assesses the chatbot's responses in terms of user-centered aspects, measuring the transparency, clarity, and comprehensibility of its decision-making process \cite{wahde2021five}. This evaluation allows users and healthcare professionals to understand the reasoning behind the chatbot's recommendations or actions. Hence, by interpretability metric, we can also evaluate the reasoning ability of chatbots which involves assessing how well a model's decision-making process can be understood and explained.
Interpretability ensures that the chatbot's behavior can be traced back to specific rules, algorithms, or data sources \cite{broniatowski2021psychological}.

\subsection{Empathy}

Empathy is the ability to understand and share the feelings of another person. Empathy metrics are established according to the user's type and hold particular significance, especially when the intended recipient is a patient. These metrics ensure that the chatbots consider end-users emotional support, trust, concerns, fairness, and health literacy \cite{zhou2020design, welivita2020taxonomy, svikhnushina2022ieval, ilicki2023framework}. 
Empathy also plays a crucial role in building trust between users and chatbots. Unempathetic responses can erode trust and credibility in the system, as users may feel unheard, misunderstood, or invalidated. 
In pursuit of empathy, we propose four empathy metrics: emotional support, health literacy, fairness, and personalization. 

The \textbf{Emotional Support} metric evaluates how chatbots incorporate user emotions and feelings.
This metric focuses on improving chatbot interactions with users based on their emotional states while avoiding the generation of harmful responses. It encompasses various aspects such as active listening, encouragement, referrals, psychoeducation, and crisis interventions \cite{meng2021emotional}.

The \textbf{Health Literacy} metric assesses the model's capability to communicate health-related information in a manner understandable to individuals with varying levels of health knowledge. This evaluation aids patients with low health knowledge in comprehending medical terminology, adhering to post-visit instructions, utilizing prescriptions appropriately, navigating healthcare systems, and understanding health-related content \cite{oniani2022toward}. For instance, ``pneumonia is hazardous" might be challenging for a general audience, while ``lung disease is dangerous" could be a more accessible option for people with diverse health knowledge. 

The \textbf{Fairness} metric evaluates the impartiality and equitable performance of healthcare chatbots. This metric assesses whether the chatbot delivers consistent quality and fairness in its responses across users from different demographic groups, considering factors such as race, gender, age, or socioeconomic status \cite{ahmad2020fairness, ahmad2021fairness}. 
Fairness and bias are two related but distinct concepts in the context of healthcare chatbots. Fairness ensures equal treatment or responses for all users, while bias examines the presence of unjustified preferences, disparities, or discrimination in the chatbot's interactions and outputs \cite{hague2019benefits, hariri2023unlocking}.
For instance, a model trained on an imbalanced dataset, with dominant samples from white males and limited samples from Hispanic females, might exhibit bias due to the imbalanced training dataset. Consequently, it may provide unfair responses to Hispanic females, as their patterns were not accurately learned during the training process. Enhancing fairness within a healthcare chatbot's responses contributes to increased reliability by ensuring that the chatbot consistently provides equitable and unbiased answers.

The \textbf{Personalization} metric gauges the degree of customization and individualization in the chatbot's conversations. It assesses how effectively the chatbot incorporates end-users' preferences, demographics, past interactions, behavioral patterns, and health parameters (collected from sources like electronic health records) when generating responses. 
Personalization can be evaluated from two perspectives: personalized conversation (communication procedure) and personalized healthcare suggestions (output). The metric, can be obtained through subjective human-based evaluation methods~\cite{cook2022measuring}. Personalization enhances the reliability of a chatbot by tailoring its interactions and healthcare recommendations to individual users, ensuring that responses align closely with their preferences and health-related data.
 
\subsection{Performance} \label{sec:performance}

Performance metrics are essential in assessing the runtime performance of healthcare conversational models, as they significantly impact the user experience during interactions. From the user's perspective, two crucial quality attributes that healthcare chatbots should primarily fulfill are \textbf{usability} and \textbf{latency}.
\textbf{Usability} refers to the overall quality of a user's experience when engaging with chatbots across various devices, such as mobile phones, desktops, and embedded systems.
\textbf{Latency} measures the round-trip response time for a chatbot to receive a user's request, generate a response, and deliver it back to the user. Low latency ensures prompt and efficient communication, enabling users to obtain timely responses. It is important to note that performance metrics may remain invariant concerning the three confounding variables (user type, domain type, and task type). In the following sections, we outline the performance metrics for healthcare conversational models.

The \textbf{Memory Efficiency} metric quantifies the amount of memory utilized by a healthcare chatbot. Popular LLMs, such as GPT-4, Llama, and BERT, often require large memory capacity \cite{brown2020language,openai2023gpt4,touvron2023llama,devlin2019bert, Lee_2019}, making it challenging to run them on devices with limited memory, such as embedded systems, laptops, and mobile phones \cite{zhuang2023survey}.

The \textbf{FLoating point OPerations (FLOP)} metric quantifies the number of floating point operations required to execute a single instance of healthcare conversational models. This metric provides valuable insights into the computational efficiency and latency of healthcare chatbots, aiding in their optimization for faster and more efficient response times.

The \textbf{Token Limit} metric evaluates the performance of chatbots, focusing on the number of tokens used in multi-turn interactions.
The number of tokens significantly impacts the word count in a query and the computational resources required during inference. As the number of tokens increases, the memory and computation needed also increase \cite{hoffmann2022an}, leading to higher latency and reduced usability. 

The \textbf{Number of Parameters} of the LLM model is a widely used metric that signifies the model's size and complexity. A higher number of parameters indicates an increased capacity for processing and learning from training data and generating output responses. 
Reducing the number of parameters, which often leads to decreased memory usage and FLOPs, is likely to improve usability and latency, making the model more efficient and effective in practical applications.

\section{Challenges in Evaluating Healthcare Chatbots} \label{sec:4}

In this section, we elucidate the challenges and pertinent factors essential for the evaluation of healthcare chatbots using the proposed user-centered metrics. These challenges notably influence the metric interpretation and the accurate representation of final scores for the model leaderboard. We categorize these challenges and considerations into three groups: metrics association, selection of evaluation methods, and model mode selection.

\subsection{Metrics Association}
The proposed metrics demonstrate both within-category and between-category associations, with the potential for negative or positive correlations among them. Within-category relations refer to the associations among metrics within the same category. For instance, 
within the accuracy metrics category, up-to-dateness and groundedness show a positive correlation, as ensuring the chatbot utilizes the most recent and valid information enhances the factual accuracy of answers, thereby increasing groundedness.

Between-category relations occur when metrics from different categories exhibit correlations. For instance, metrics in trustworthiness and empathy may be correlated.
Empathy often necessitates personalization, which can potentially compromise privacy and lead to biased responses.

A significant relationship exists between performance metrics and the other three categories. 
For instance, the number of parameters in a language model can impact accuracy, trustworthiness, and empathy metrics. An increase in parameters may introduce complexity, potentially affecting these metrics positively or negatively. Conversely, a low parameter count can limit the model's knowledge acquisition and influence the values of these metrics.

\subsection{Evaluation Methods}
Various automatic and human-based evaluation methods can quantify each metric, and the selection of evaluation methods significantly impacts metric scores. 
Automatic approaches utilize established benchmarks to assess the chatbot's adherence to specified guidelines, such as using robustness benchmarks alongside metrics like ROUGE or BLEU to evaluate model robustness.

However, a notable concern arises when employing existing benchmarks (see Table \ref{tab:healthmetrics}) to automatically evaluate relevant metrics. These benchmarks may lack comprehensive assessments of the chatbot model's robustness concerning confounding variables specific to the target user type, domain type, and task type. Ensuring a thorough evaluation of robustness requires diverse benchmarks that cover various aspects of the confounding variables.

Human-based methods involve providing questions or guidelines to human annotators who score the chatbot's generated answers based on given criteria. This approach presents two main challenges: subjectivity and the need for a variety of domain expert annotators. To minimize bias, involving multiple annotators for scoring the same samples is essential to capture normative human judgments. Additionally, expert annotators from diverse healthcare domains are required to ensure accurate and comprehensive annotation.

It is crucial to acknowledge two strategies for scoring metrics. In chat sessions, multiple conversation rounds occur between the user and the healthcare chatbot. The first strategy involves scoring after each individual query is answered (per answer), while the second strategy involves scoring the healthcare chatbot once the entire session is completed (per session). Some metrics, like intrinsic ones, perform better when assessed on a per-answer basis \cite{nistTextREtrieval}.

 \subsection{Model Prompt Techniques and Parameters}
Prompt engineering \cite{zhou2022large} significantly impacts the responses generated by healthcare chatbots, and the choice of prompt technique plays a pivotal role in achieving improved answers. Various prompting methods, such as zero-shot, few-shot, chain of thought generated with evidence, and persona-based approaches, have been proposed in the literature. 

Apart from prompting techniques, evaluation based on model parameters during inference is also crucial. Modifying these parameters can influence the chatbot's behavior when responding to queries. For example, adjusting the beam search parameter \cite{ge2023openagi} can impact the safety level of the chatbot's answers, and similar effects apply to other model parameters like temperature \cite{chung-etal-2023-increasing}, which can influence specific metric scores.

\section{Toward an Effective Evaluation Framework}
\begin{figure}[!t]
\includegraphics[width=\linewidth, trim={15.5cm 4.5cm 15cm 3.5cm},clip]{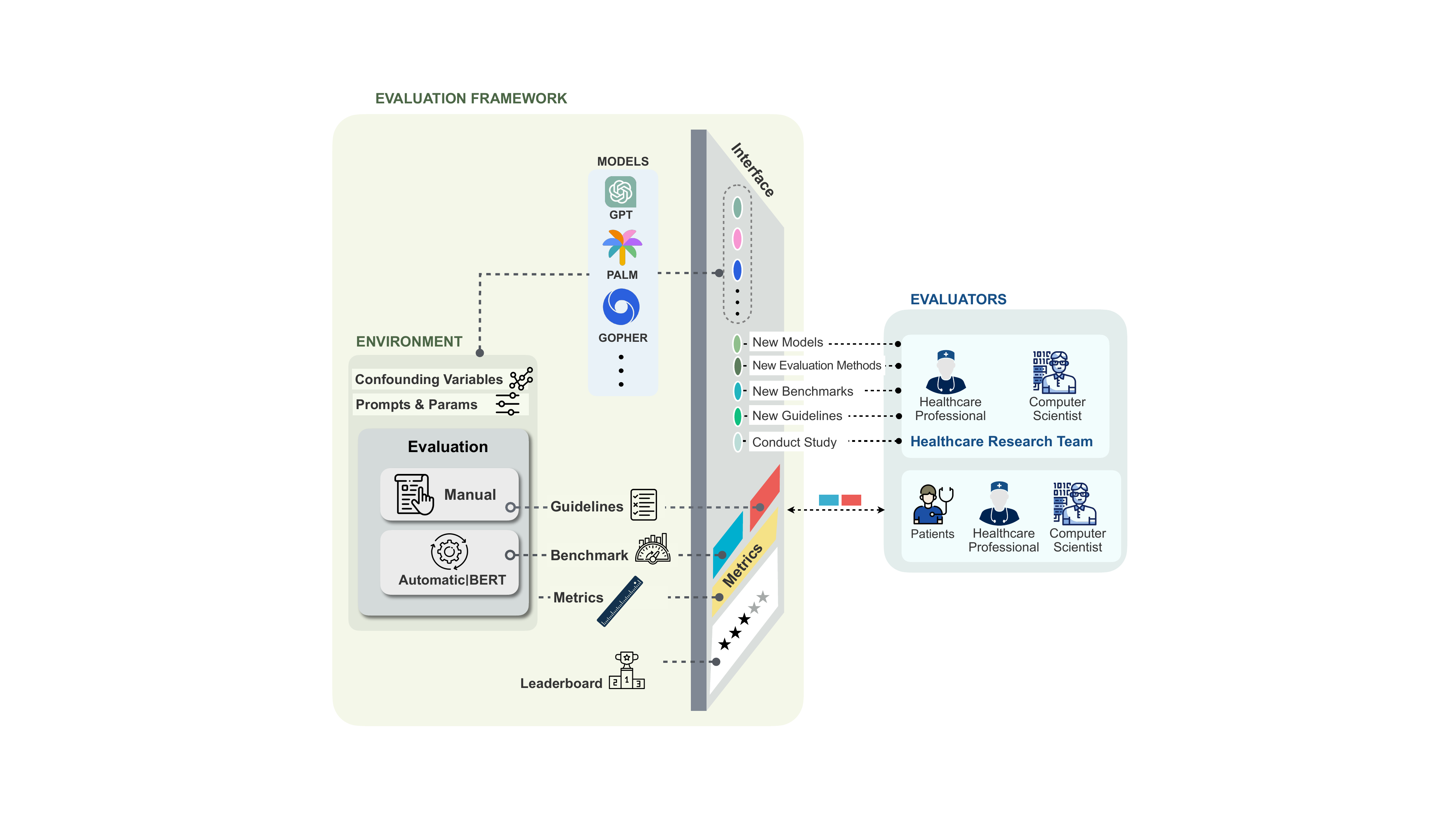}
\centering
\caption{\textbf{An illustrative high-level representation of an evaluation framework} containing five main components: models, environment, interface, interacting users, and leaderboard.} {\label{fig:evalframework}}
\end{figure}

Considering the aforementioned deliberations regarding the requirements and complexities entailed in the evaluation of healthcare chatbots, it is of paramount importance to institute effective evaluation frameworks. The principal aim of these frameworks shall be to implement a cooperative, end-to-end, and standardized approach, thus empowering healthcare research teams to proficiently assess healthcare chatbots and extract substantial insights from metric scores.

In this context, Figure \ref{fig:evalframework} presents an illustrative high-level representation of such an evaluation framework.This framework is intended to act as the foundational codebase for future benchmarks and guidelines. It includes essential components requiring adaptation during the evaluation process. Notably, while recent studies \cite{ahuja2023mega, ilicki2023framework, liu2023gpteval, reddy2023evaluating} have introduced various evaluation frameworks, it is important to recognize that these may not fully cater to the specific needs of healthcare chatbots. Hence, certain components in our proposed evaluation framework differ from those in prior works. In the ensuing sections, we expound on these components and discuss the challenges that necessitate careful consideration and resolution.

The term \textbf{Models} within the evaluation framework pertains to both current and prospective healthcare chatbot models. The framework should enable seamless interaction with these models to facilitate efficient evaluation.

The evaluation framework encompasses the configurable \textbf{Environment}, where researchers establish specific configurations aligned with their research objectives. The three key configuration components consist of confounding variables, prompt techniques and parameters, and evaluation methods.
\begin{enumerate}
    \item The \textbf{Confounding Variables} component is pivotal, as it stores configurations related to users, domains, and task types. The ability to adjust these variables in the evaluation framework ensures alignment among all stakeholders evaluating the target healthcare chatbot model, fostering a consistent and uniform evaluation perspective.

   \item  The \textbf{Prompt Techniques and Parameters} component enables the configuration of desired prompting techniques and LLM parameters. Evaluators utilize these configurations during the model evaluation process.

    \item The \textbf{Evaluation} component represents a critical aspect of the evaluation framework, providing essential tools for evaluators to calculate individual metric scores, category-level metric scores, and a comprehensive total score for the desired healthcare chatbot model. Figure \ref{fig:evalframework} illustrates the tools required in this component. To create a comprehensive evaluation process, specific requirements must be addressed. These include developing tailored benchmarks for healthcare domains, establishing detailed guidelines for human-based evaluations, introducing innovative evaluation methods designed explicitly for healthcare metrics, and providing evaluation tools to support annotators. 
\end{enumerate}

One primary requirement for a comprehensive evaluation component is the development of healthcare-specific benchmarks that align with identified metric categories similar to introduced benchmarks in Table \ref{tab:healthmetrics} but more concentrated on healthcare. These benchmarks should be well-defined, covering each metric category and its sub-groups to ensure thorough testing of the target metrics. Tailored benchmarks for specific healthcare users, domains, and task types should also be established to assess chatbot performance within these confounding variables. When combined with automatic evaluation methods like ROUGE and BLEU, these benchmarks enable scoring of introduced extrinsic metrics.

The second crucial requirement involves creating comprehensive human guidelines for evaluating healthcare chatbots with the aid of human evaluators. These guidelines facilitate manual scoring of metrics. Healthcare professionals can assess the chatbot's performance from the perspective of the final users, while intended users, such as patients, can provide feedback based on the relevance and helpfulness of answers to their specific questions and goals. As such, these guidelines should accommodate the different perspectives of the chatbot's target user types.

To ensure objectivity and reduce human bias, providing precise guidelines for assigning scores to different metric categories is indispensable. This fosters consistency in scoring ranges and promotes standardized evaluation practices. Utilizing predefined questions for evaluators to assess generated answers has proven effective in improving the evaluation process. By establishing standardized questions for each metric category and its sub-metrics, evaluators exhibit more uniform scoring behavior, leading to enhanced evaluation outcomes \cite{thoppilan2022lamda, glaese2022improving}.

The third crucial requirement involves devising novel evaluation methods tailored to the healthcare domain. These methods should integrate elements from the previous requirements, combining benchmark-based evaluations with supervised approaches to generate a unified final score encompassing all metric categories. Moreover, the final score should account for the assigned priorities to each metric category. For example, if trustworthiness outweighs accuracy in a specific task, the final score should reflect this prioritization.
    
The integration of the aforementioned requirements should result in the desired scores, treating the evaluation component as a black box. Nevertheless, an unexplored avenue lies in leveraging BERT-based models, trained on healthcare-specific categorization and scoring tasks. By utilizing such models, it becomes possible to calculate scores for individual metrics, thereby augmenting the evaluation process. 

To facilitate effective evaluation and comparison of diverse healthcare chatbot models, the healthcare research team must meticulously consider all introduced configurable environments. 
By collectively addressing these factors, the interpretation of metric scores can be standardized, thereby mitigating confusion when comparing the performance of various models.

The \textbf{Interface} component serves as the interaction point between the environment and users. Through this interface, interacting users can configure the environment by selecting the desired model for interaction, modifying model parameters, choosing the target user type, accessing evaluation guidelines, selecting the evaluation method, utilizing the latest introduced benchmarks, and more. Furthermore, the interface enables researchers to create new models, evaluation methods, guidelines, and benchmarks within the provided environment. 

The \textbf{Interacting users} of the evaluation framework serve different purposes and can be categorized into two main groups: evaluators and healthcare research teams.
Evaluators utilize the evaluation framework through the interface to assess healthcare chatbot models and score the metrics.
Healthcare research teams encompass computer and data scientists who contribute to new model creation and the development of novel evaluation methods. Additionally, it includes healthcare professionals who conduct new studies or contribute to the establishment of new benchmarks and guidelines. For instance, a healthcare research team might evaluate the performance of ChatGPT in answering mental health queries. In this scenario, healthcare professionals can introduce a new benchmark in the evaluation framework or provide novel guidelines to evaluators for evaluating ChatGPT based on metrics and assigning scores. Alternatively, the healthcare research team can use the existing evaluation tools to evaluate ChatGPT's performance in mental health. Eventually, the healthcare research team can report their findings and scores obtained through the evaluation process.

The \textbf{Leaderboard} represents the final component of the evaluation framework, providing interacting users with the ability to rank and compare diverse healthcare chatbot models. It offers various filtering strategies, allowing users to rank models according to specific criteria. For example, users can prioritize accuracy scores to identify the healthcare chatbot model with the highest accuracy in providing answers to healthcare questions. Additionally, the leaderboard allows users to filter results based on confounding variables, facilitating the identification of the most relevant chatbot models for their research study.

\section {Conclusion}

Generative AI, particularly chatbots, shows great potential in revolutionizing the healthcare industry by offering personalized, efficient, and proactive patient care. This paper delved into the significance of tailored evaluation metrics specifically for healthcare chatbots. We introduced a comprehensive set of user-centered evaluation metrics, grouped into four categories: accuracy, trustworthiness, empathy, and computing performance. The study highlighted the potential impact of confounding variables on metric definition and evaluation. Additionally, we emphasized how these metrics can address pertinent issues and enhance the reliability and quality of healthcare chatbot systems, ultimately leading to an improved patient experience. Lastly, we examined the challenges associated with developing and implementing these metrics in the evaluation process. 

Future directions for this work involve the implementation of the proposed evaluation framework to conduct an extensive assessment of metrics using benchmarks and case studies. We aim to establish unified benchmarks specifically tailored for evaluating healthcare chatbots based on the proposed metrics. Additionally, we plan to execute a series of case studies across various medical fields, such as mental and physical health, considering the unique challenges of each domain and the diverse parameters outlined in Section \textit{Evaluation Methods}.

\section{Competing Interests}
Y.W. is a collaborator of HealthUnity, consults for Pfizer Inc., and has ownership/equity interests in BonafideNLP, LLC. D.O. is a collaborator of HealthUnity. All other authors declare no Competing Financial or Non-Financial Interests.

\section{Author Contributions}
MA and EKH conducted the research, analyzed the findings, and drafted the manuscript. MA and EKH are co-first authors. IA played a key role in designing the study and revised the paper critically. DO contributed to drafting performance sub-section and revised the paper. ZSHA contributed to give guidance, revise critically the paper, and the design of the visualizations. AT and BL revised and validated the study from clinical perspectives. RS refined the paper and ensured alignment with NIST metrics. ZY contributed to drafting one proposed metric. YW and OG participated in the revising process. LJL, RJ, and AMR led the study, did mentoring, provided guidance throughout, and conducted critical revisions of the manuscript. All authors read and approved the final manuscript.

\section{Disclaimer}

Certain commercial systems are identified in this paper. Such identification does not imply recommendation or endorsement by NIST; nor does it imply that the products identified are necessarily the best available for the purpose. Further, any opinions, findings, conclusions or recommendations expressed in this material are those of the authors and do not necessarily reflect the views of NIST, other supporting U.S. government or corporate organizations.

\section{Acknowledgments}
We would like to thank the following NIST people for their in-depth comments:  Ian Soboroff, Hoa Dang, Jacob Collard, and Reva Schwartz. Furthermore, we express our gratitude to Nigam Shah from Stanford for his valuable feedback, which has contributed to the enhancement of the paper.

\bibliographystyle{naturemag}
{\footnotesize
\bibliography{ref_submit.bib}}

\end{document}